\documentclass[conference]{IEEEtran}
\IEEEoverridecommandlockouts
\usepackage{cite}
\usepackage{amsmath,amssymb,amsfonts}
\usepackage{algorithmic}
\usepackage{graphicx}
\usepackage{textcomp}
\usepackage{xcolor}
\usepackage{hyperref}
\usepackage{fontawesome}
\usepackage{bbding} 
\usepackage{booktabs} 
\def\BibTeX{{\rm B\kern-.05em{\sc i\kern-.025em b}\kern-.08em
    T\kern-.1667em\lower.7ex\hbox{E}\kern-.125emX}}
    
\begin{document}
\title{Seismic Fault SAM: Adapting SAM with Lightweight Modules and 2.5D Strategy for Fault Detection\\
\author{
    \IEEEauthorblockN{
        Ran Chen\textsuperscript{*}, 
        Zeren Zhang\textsuperscript{*}, 
        Jinwen Ma\textsuperscript{\Envelope}
    }
    \IEEEauthorblockA{
        \textit{Department of Information and Computational Sciences} \\
        \textit{School of Mathematical Sciences, Peking University}\\
        Beijing 100871, China \\
        \{chenran, eric\_zhang\}@stu.pku.edu.cn, jwma@math.pku.edu.cn
    }
    \thanks{\textsuperscript{*} These authors contributed equally to this work.}
    \thanks{\textsuperscript{\Envelope} Corresponding author: Jinwen Ma (jwma@math.pku.edu.cn)}
}
\thanks{This work was supported by the Natural Science Foundation of China under grant 62071171.}
}

\maketitle

\begin{abstract}
Seismic fault detection holds significant geographical and practical application value, aiding experts in subsurface structure interpretation and resource exploration. Despite some progress made by automated methods based on deep learning, research in the seismic domain faces significant challenges, particularly because it is difficult to obtain high-quality, large-scale, open-source, and diverse datasets, which hinders the development of general foundation models. Therefore, this paper proposes Seismic Fault SAM, which, for the first time, applies the general pre-training foundation model—Segment Anything Model (SAM)—to seismic fault interpretation. This method aligns the universal knowledge learned from a vast amount of images with the seismic domain tasks through an Adapter design. Specifically, our innovative points include designing lightweight Adapter modules, freezing most of the pre-training weights, and only updating a small number of parameters to allow the model to converge quickly and effectively learn fault features; combining 2.5D input strategy to capture 3D spatial patterns with 2D models; integrating geological constraints into the model through prior-based data augmentation techniques to enhance the model's generalization capability. Experimental results on the largest publicly available seismic dataset, Thebe, show that our method surpasses existing 3D models on both OIS and ODS metrics, achieving state-of-the-art performance and providing an effective extension scheme for other seismic domain downstream tasks that lack labeled data.

\end{abstract}


\section{Introduction}

Seismic fault detection involves processing and interpreting seismic data to recognize the structure and distribution of subsurface faults. This task is crucial in geophysics and geology, playing a significant role in petroleum exploration and geological risk assessment \cite{hale2012fault}. However, due to the inherent uncertainty in geological structures and the interference during seismic data acquisition and processing, formation features and noise are often mistaken for reflection discontinuities (i.e., faults), increasing the difficulty of identification \cite{hale2012fault}. Additionally, manual fault detection is labor-intensive and susceptible to interpreter bias \cite{admasu2006autotracking}. Traditional machine learning methods based on feature design and extraction struggle to capture high-level features. They are susceptible to noise and require heuristic parameter settings, which limit their generalization capability \cite{an2021deep}.

In recent years, deep learning algorithms based on CNN architectures have gradually gained widespread application in fault detection \cite{wu2019faultseg3d, an2021deep, an2023current}. However, these methods rely on a large amount of labeled data, and obtaining seismic data is challenging. The 3D labeling process is time-consuming and expensive, with limited resources available for open annotation. This limitation results in models being constrained by synthetic datasets and exhibiting poor generalization performance on field datasets \cite{an2023understanding}. To address this challenge, researchers have proposed transfer, semi-supervised, and self-supervised learning methods \cite{an2023understanding, an2023current, dou2023faultssl, zhang2024improving}. However, it's still challenging to get high-quality, large-scale, open-source, and diverse datasets for pre-training general foundational models \cite{an2023current}.
 
 To get rid of this difficulty, we explore whether a large computer vision model pre-trained on natural images, the Segment Anything Model (SAM) \cite{kirillov2023segment}, can be applied to seismic fault detection. SAM is trained on a large dataset containing 11 million images and over 1 billion masks, demonstrating its powerful segmentation capabilities across various scenarios. However, our experiments show that when SAM is used for fault detection tasks, its performance is limited by the significant differences between natural and seismic images. SAM is designed only for 2D images, whereas 3D fault detection models can yield more robust and continuous results \cite{zhang2024improving}. At the same time, most current methods directly analyze the input seismic data while neglecting known geological constraints. Research is still needed to incorporate this prior knowledge into deep learning models \cite{an2023current}. 

This paper explores the potential advantages of pre-training on natural images for seismic fault detection. It experimentally compares the performance of seismic image and natural image pre-training methods on segmentation models with Transformer architectures. The results indicate that under the current limited availability of seismic image datasets, models pre-trained on natural images perform better in downstream fault detection tasks, significantly improving model convergence speed and effectiveness. To fully leverage SAM's powerful feature extraction capabilities pre-trained on natural images, we propose Seismic Fault SAM: the first model that effectively adapted SAM for fault detection using Adapters.

Specifically, we design a lightweight Adapter module, allowing SAM to efficiently learn features pertinent to fault detection by freezing the original pre-training network weights and updating only a few parameters. Additionally, we introduce a 2.5D fault detection strategy that integrates spatiotemporal features and achieves 3D fault detection using 2D SAM. Third, considering the layered structure of seismic images and the prior knowledge that seismic waves may deform and shift as they propagate through different geological conditions, we incorporate characteristic information into field seismic data through data augmentation to increase its diversity. Experiments demonstrate that our method achieves state-of-the-art performance on the largest publicly available field dataset, Thebe \cite{an2021deep}, with optimal image scale (OIS) and optimal dataset scale (ODS) metrics reaching 0.881 and 0.879, respectively.

Therefore, the contributions of this paper are as follows:

\begin{itemize}
    \item We propose a lightweight adaptation framework for fault detection that enables efficient learning and rapid updates without retraining. This framework allows the SAM model to be effectively transferred to seismic tasks without relying on extensive field seismic data, thus avoiding pre-training on millions of parameters.
    
    \item We incorporate domain-specific prior knowledge of seismic characteristics and perform specific transformation operations through data augmentation. This approach improves the model's understanding of the layered structure of seismic images, making it more robust to changes in fault direction and various deformations, thereby enhancing the model's generalization ability to diverse data.

    \item We introduce adjacent seismic slice information to construct a 2.5D model and utilize 2D SAM combined with 3D data for training, which enhances prediction continuity and stability. This method enables the model to possess 3D spatial understanding capabilities, ultimately achieving predictive performance that surpasses state-of-the-art 3D models.
    
\end{itemize}

\section{Related Work}

In the field of automatic fault detection, seismic attribute methods such as coherence \cite{marfurt19983}, anisotropy \cite{ruger1997using}, and machine learning methods \cite{di2019improving, an2023current} have gradually been proposed. In recent years, deep learning has autonomously learned and extracted features from seismic images, reducing the reliance on feature engineering. Wu et al. \cite{wu2019faultseg3d} proposed a method for constructing synthetic datasets and trained an end-to-end 3D-Unet network named FaultSeg3D. An et al. \cite{an2021deep} released the largest publicly available field fault dataset, Thebe, and achieved good results using the 2D segmentation network Mobile DeepLabV3+. Yang et al. \cite{yang2023multi} proposed a dual-task deep learning strategy that achieves parameter sharing between horizon extraction and fault detection branches. Most of these works are based on CNNs, but Transformer-based methods \cite{vaswani2017attention} have shown excellent performance in capturing long-range dependencies and global information. For example, Bomfim et al. \cite{bomfim2023transformer} achieved better results than traditional CNNs on Brazilian Pre-salt Seismic Data using the TransUNet network, combining CNN and Transformer. Tang et al. \cite{tang2023fault} used a 2.5D Transformer U-net method to predict more detailed faults. However, these supervised methods are limited by the quantity and quality of labeled data. The semi-supervised work FaultSSL \cite{dou2024faultssl} is based on a mean teacher structure, integrating supervised and self-supervised learning. FaultSeg Swin-UNETR \cite{zhang2024improving} makes improvements to the Swin-UNETR architecture to enhance multi-scale decoding and fusion capabilities for seismic fault detection.

ViT (Vision Transformer) \cite{dosovitskiy2020image} has demonstrated outstanding performance in computer vision tasks. Swin Transformer \cite{liu2021swin}, built on ViT, introduces hierarchical feature representation and a sliding window strategy. SAM is a recently developed vision foundation model for prompt-based image segmentation that has been successfully applied in medical fields \cite{chen2023ma}, but its transferability to the seismic domain has yet to be fully explored. Although SFM \cite{sheng2023seismic} uses self-supervised learning to pre-train a Transformer-based model for tasks like interpolation, denoising, and inversion, it doesn't address fault detection tasks.

\section{Method}

\begin{figure*}[t]
    \centering
     \includegraphics[width = 17cm]{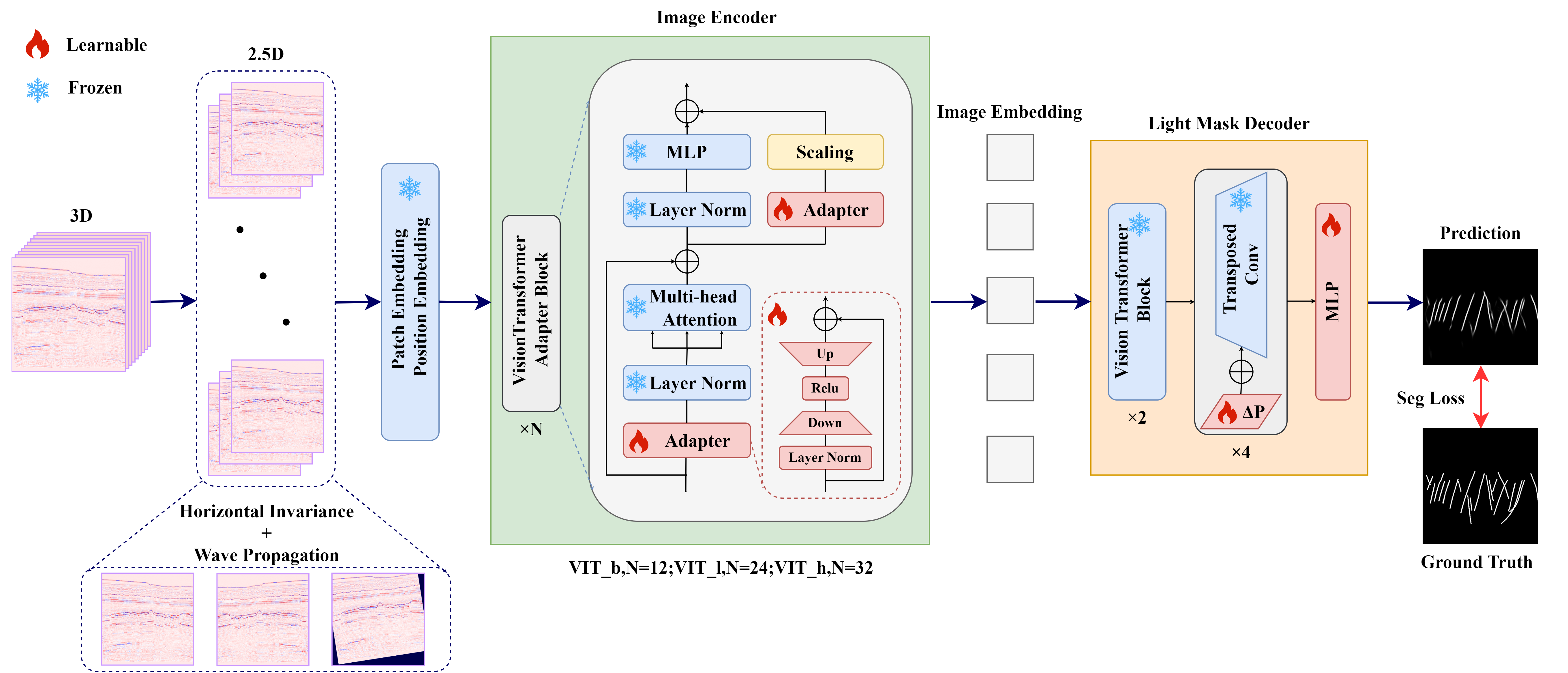}
    \caption{The overview of our proposed 2.5D SAM adaptation framework for Seismic Fault Detection (Seismic Fault SAM).}
    \label{Mainfig}
\end{figure*}

This section will introduce how to adapt the original SAM architecture for seismic fault segmentation tasks. Fig.~\ref{Mainfig} shows an overview of our approach. First, we briefly introduce the SAM architecture. We provide a detailed explanation of the 2.5D strategy, the data augmentation strategy based on seismic priors, and the technical details of the adaptation image encoder and the lightweight mask decoder.

\subsection{Overview of SAM}

SAM consists of three main components: an image encoder, a prompt encoder, and a mask decoder. The image encoder uses a ViT pre-trained with MAE, which segments the image into patches and transforms them into 16$\times$ downsampled embeddings through a series of Vision Transformer blocks. The prompt encoder handles sparse (e.g., points, boxes) and dense (e.g., masks) prompts, allowing the model to respond flexibly to different types. The lightweight mask decoder is an improved Transformer decoder block that computes cross-attention between image and prompt embeddings and generates segmentation masks through transposed convolution layers and MLP. Due to the domain differences between natural and seismic images, the model requires specific adaptation and fine-tuning. 

\subsection{2.5D Strategy}\label{AA}

SAM is initially pre-trained on 2D images, but seismic images contain critical 3D spatial information. To effectively adapt 2D SAM for seismic images, it is essential to introduce critical third-dimensional information during fine-tuning. Therefore, we adopt a 2.5D approach to integrate correlations between slices, extracting spatial insights necessary for fault detection and bridging the complexity of seismic data with the 2D pre-trained model. Specifically, to make 3D images compatible with the 2D SAM, we extract a set of adjacent slices $\textbf{x}$ before feeding the images into the SAM backbone, 

\begin{equation}
\textbf{x} = \left\{x_{i-\frac{M-1}{2}},...,x_i,...,x_{i+\frac{M-1}{2}} \right\}_{i=1}^B,\textbf{x} \in \mathbb{R}^{B \times M \times H \times W}
\end{equation}

Here, $B$ denotes
the batch size, $M$ denotes the number of adjacent slices,
and $H \times W$ denotes the slice shape. We merge the adjacent slices into the channel dimension to form a feature map with a channel number of $M$ as the 2.5D data input, and the corresponding output is the fault prediction of the central slice. This approach enables the model to capture the spatial correlations between slices during patch-by-patch prediction. The model compensates for the 3D information through this additional fusion structure while retaining as much pre-trained weight as possible, effectively extracting the spatiotemporal relationships of seismic data.

\begin{figure}[htbp]
\centerline{\includegraphics[width=9cm]{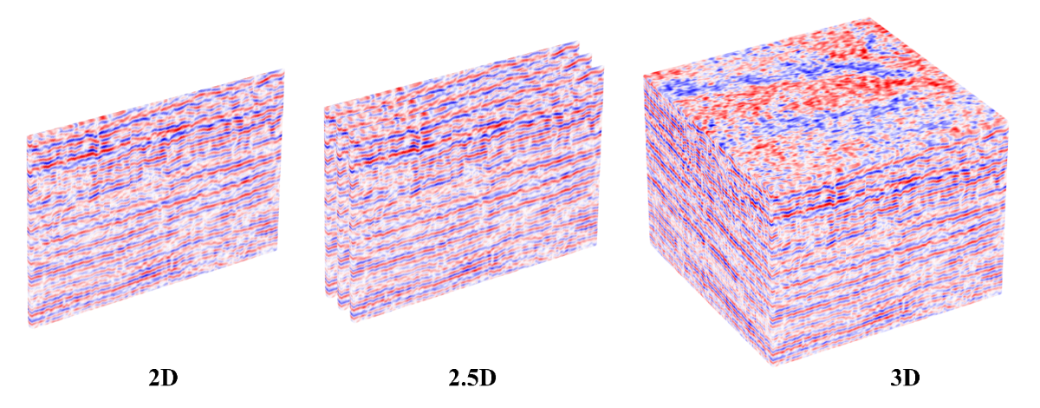}}
\caption{Illustration of the inputs to 2D, 2.5D, and 3D segmentation models.}
\label{25D}
\end{figure}

\subsection{Data Augmentation}\label{BB}

To address the limitations of the scale of field seismic datasets and the poor generalization performance of models limited to a specific field, we leverage prior knowledge from the seismic domain to bridge the gap between SAM's pre-trained data and field seismic data. This approach enhances SAM's fault characterization capabilities while retaining its powerful segmentation abilities. Data augmentation is a common method for improving the performance of deep learning models by introducing prior information. Specifically, it enhances the diversity of the training set by transforming images from the original dataset using the most effective augmentation strategies selected based on domain knowledge or expert insights.

Seismic images typically exhibit horizontal invariance \cite{marfurt19983}, as seismic data usually present layered structures that show symmetry in different directions. For instance, the reflection layers in seismic data exhibit similar characteristics when seismic waves propagate along different paths. Fig.~\ref{Mainfig} (left) demonstrates how horizontal flipping can simulate the symmetrical changes of seismic reflection layers in different directions, aiding the model in better understanding and adapting to these layered structures. This enhances the model's robustness to changes in fault and stratification directions. Furthermore, during the propagation of seismic waves, deformations such as rotation, scaling, and translation occur due to the heterogeneity of subsurface media and geological complexities \cite{ruger1997using}. Using random affine transformations to simulate these features can help the model learn the deformation characteristics of seismic waves under different conditions. This improves the model's generalization ability in complex geological conditions, making it more reliable in identifying faults across data collected under varying acquisition conditions.

These data augmentation techniques effectively incorporate prior knowledge from the seismic domain, enhancing the model's robustness and generalization capabilities. Our approach integrates the expertise of geologists into the training dataset, expanding the diversity of seismic data and reducing the cumbersome task of fully annotating large volumes of seismic data.

\subsection{Adapter}\label{CC}

Adapter is a very simple yet effective foundational model transfer technology that usually requires learning only a small number of parameters (typically less than 5$\%$ of the total) while keeping most parameters frozen. This approach can achieve better results than full fine-tuning because it avoids catastrophic forgetting and retains the general knowledge needed for segmentation. Therefore, this paper proposes an Adapter framework specifically designed for fault detection. Designing lightweight modules injects specific guiding information from relatively small amounts of data to align SAM with the fault detection task, requiring less fine-tuning time and resource consumption.

\subsubsection{Image Encoder}

The image encoder of SAM is a massive component in the model, contrasting with the lightweight mask decoder. These parameters incorporate the general knowledge needed for strong segmentation capabilities and, therefore, need to be retained and enhanced for reusability. To achieve this, we design two bottleneck Adapters in each standard ViT block and integrate them into specified locations for efficient fine-tuning, as shown in the green parts of Fig.~\ref{Mainfig}.

Specifically, the bottleneck model sequentially includes Layer Norm, down-projection, ReLU activation, and up-projection, which can be represented as 

\begin{equation}
   Apater (\mathbf{X})=\mathbf{X}+\sigma\left( Norm (\mathbf{X}) W_{down}\right) W_{up} 
\end{equation}
where $\mathbf{X}$ denoting feature maps, $W_{\text {down }} \in \mathbb{R}^{c \times d}$ and $W_{u p} \in$ $\mathbb{R}^{d \times c}$ indicate the down-projection layer and up-projection layer, respectively, and $\sigma(\cdot)$ is the activation function, such as Relu. The purpose of down-projection is to compress the original dimensional features into a lower dimensional representation to control the number of newly introduced parameters. At the same time, the up-projection restores the original dimensions. In the standard ViT block, the first Adapter is placed before the Norm layer, and the second Adapter is placed in the skip connection of the residual path of the MLP layer after multi-head attention, scaled with a balance factor.

\subsubsection{Light Mask Decoder}

We emphasize lightweight design, progressive upsampling, and full automation for the mask decoder (as shown in the orange parts of Fig.~\ref{Mainfig}). The mask decoder in the original SAM comprises two Transformer blocks, two transposed convolution layers, and one MLP layer. Each 16 $\times$ 16 patch is embedded into a feature vector during the image encoding process, resulting in a downsampling factor of 16 for the image embedding. The mask decoder uses two consecutive transposed convolution layers to upsample the feature map 4 times, making the final mask prediction resolution 4 times lower than the original image. This is effective for natural images where objects are typically larger and boundaries are clear, but it is unsuitable for fault detection. To better recognize small and blurry boundaries in faults, we add two more transposed convolution layers, with each layer progressively upsampling the feature map by 2 times, gradually restoring the high resolution of the original input. 

Additionally, we remove the prompt embedding part of SAM because interactive segmentation requires experts to provide accurate prompts, examine results at each step, and interact. If the prompts are inaccurate, the model is prone to failure. We aim for the model to be more intelligent, support non-expert users, and even eliminate the need for expert prompts. Therefore, we abandon the original SAM's interactive training and inference mode, opting for fully automated training and inference, enabling SAM to segment faults accurately without prompts.

Regarding the training strategy, we only need to introduce small-scale and easily adjustable incremental parameters into the convolution layers, which is an efficient fine-tuning strategy: freeze other parts of the decoder, restrict the parameters in the convolution layers to change only by $\Delta \mathbf{P}$ and add regularization on $\Delta \mathbf{P}$ to the loss function $\mathcal{L}$, as shown below, allowing  $\Delta \mathbf{P}$ to be optimized within minor variations.

\begin{equation}
\mathcal{L}_p=\mathcal{L}+\lambda_p\left(\sum_{m, n}\left|\Delta \mathbf{P}_n^m\right|_2^2\right)
\end{equation}
where $\mathbf{P}_n^m $ denotes the $m$th parameter of the $n$th element in $\mathbf{P}$, $\lambda_p$ is the weight coefficient.

With the aforementioned improvements, we transforme SAM into a potentially powerful tool for seismic fault detection.

\section{Experimental Results}

\subsection{Dataset}

In An et al.'s 2023 review \cite{an2023current}, a total of 73 seismic datasets were summarized, among which only Thebe \cite{an2021deep} collected by An's team is an openly available annotated 3D field dataset. Thebe, the largest publicly available seismic fault detection dataset, provides a critical public benchmark for research. This dataset is obtained by experts from the Fault Analysis Group at University College Dublin and contains many detailed, pixel-level expert annotations. According to the method proposed by An et al., the dataset is divided into training, validation, and test sets, which consist of the first 900 crosslines, the next 200 crosslines, and the last 703 crosslines, respectively.

\subsection{Metrics}

We follow the evaluation methods used in \cite{zhang2024improving}, using two F1 score-based evaluation metrics: optimal image scale (OIS) and optimal dataset scale (ODS). OIS calculates the best threshold for each image in the dataset, averages the corresponding F1 values, and measures the model's best performance on each image. ODS finds a global optimal threshold across the entire dataset, computes the average F1 score at this threshold, and assesses the model's performance on the whole dataset. Together, these metrics comprehensively evaluate the model's local and global performance.

\subsection{Results}
In this study, we primarily base our implementation and improvements on the official implementations. All experiments are conducted on the PyTorch platform and are trained and tested on a single NVIDIA A6000 GPU. We utilize the default settings to re-implement the comparison methods.

\begin{table}[htbp]
\caption{Comparison results of various methods on the Thebe dataset.}
\label{mainexpr}
\centering
\setlength{\tabcolsep}{1.8pt} 
\begin{tabular}{ccccc}
    \toprule
    \textbf{Model} & \textbf{Dimension} & \textbf{Pre-trained method} & \textbf{OIS} & \textbf{ODS}\\
    \midrule
    FaultSeg3D & 3D & w/o & 0.840 & 0.836 \\
    Swin-UNETR & 3D & SimMIM & 0.872 & 0.868 \\
    FaultSeg Swin-UNETR & 3D & SimMIM & 0.875 & 0.870 \\
    UNet & 2D & w/o & 0.769 & 0.766 \\
    HED & 2D & w/o & 0.811 & 0.806 \\
    RCF & 2D & w/o & 0.806 & 0.800 \\
    DeeplabM & 2D & w/o & 0.759 & 0.756 \\
    DeeplabV3 & 2D & Imagenet pre-trained & 0.849 & 0.845 \\
    ViT & 2D & w/o & 0.801 & 0.805 \\
    ViT & 2D & MAE & 0.806 & 0.801 \\
    ViT & 2D & Imagenet pre-trained & 0.848 & 0.844 \\
    Swin Transformer & 2D & w/o & 0.848 & 0.845 \\
    Swin Transformer & 2D & SimMIM & 0.857 & 0.853 \\
    Swin Transformer & 2D & Imagenet pre-trained & 0.865 & 0.862 \\
    SAM-b & 2D & SAM pre-trained & 0.187 & 0.167 \\
    SAM-b+Adapter & 2D & SAM pre-trained & 0.865 & 0.861 \\
    2.5D SAM-b+Adapter & 2.5D & SAM pre-trained & 0.870 & 0.868 \\
    2.5D SAM-h+Adapter & 2.5D & SAM pre-trained & 0.879 & 0.876 \\
    \textbf{2.5D SAM-h+Adapter+DA} & \textbf{2.5D} & \textbf{SAM pre-trained} & \textbf{0.881}
 & \textbf{0.879} \\
    \bottomrule
    \multicolumn{5}{l}{$^{\mathrm{a}}$w/o indicates without pre-training.} \\
    \multicolumn{5}{l}{$^{\mathrm{b}}$SAM-b indicates SAM-base, SAM-h indicates SAM-huge.} \\
    \multicolumn{5}{l}{$^{\mathrm{c}}$DA indicates data augmentation.}
\end{tabular}
\end{table}

As shown in Table~\ref{mainexpr}, we compare the performance of various models on the Thebe dataset and discuss the impact of different pre-training methods on model performance. In the table, SimMIM and MAE represent pre-training with corresponding methods on many unlabeled private field seismic data, consistent with the setting in \cite{zhang2024improving}. Imagenet pre-trained uses Imagenet data for pre-training, and SAM pre-trained follows the SAM pre-training method. The training epochs are set to 300, and the number of adjacent slices is set to 5.

The experimental results show that whether ViT or Swin Transformer, ImageNet pre-trained models are better than seismic dataset pre-trained models. This is because the seismic datasets currently used for pre-training are far inferior to ImageNet in scale and quality, and the strong segmentation capabilities on natural images help the model identify faults. When SAM is directly applied to the Thebe dataset, the effect is unsatisfactory due to the domain difference between natural and seismic images. However, after introducing the Adapter framework on SAM-b, only 2$\%$ of the parameters need to be updated, which surpasses Swin Transformer. After presenting the 2.5D input method, the performance is further improved, approaching the effect of advanced 3D models. After using SAM-h, which contains more general segmentation knowledge, as the backbone network, the model performance surpasses the 3D SOTA model FaultSeg Swin-UNETR. Combining the data enhancement strategy based on the domain prior, the model achieved the most advanced effects, with OIS and ODS reaching 0.881 and 0.879, respectively.

\begin{figure}[htbp]
\centerline{\includegraphics[width=9cm]{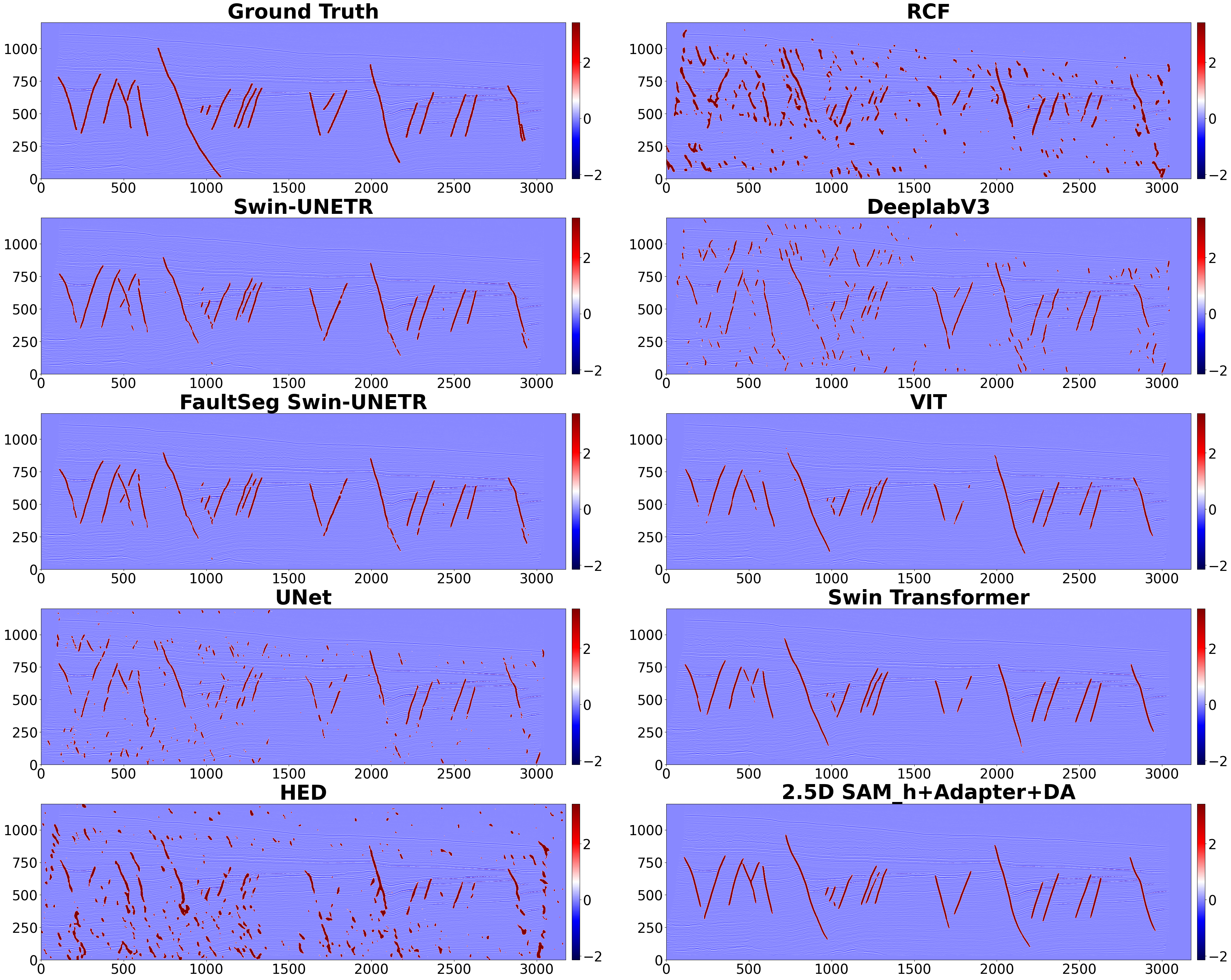}}
\caption{Fault prediction results on the 150th crossline slice of the Thebe test set with a threshold of 0.7}
\label{2Dshow}
\end{figure}

Fig.~\ref{2Dshow} shows the fault prediction results of different models, with the red area indicating the parts where the fault prediction probability exceeds 70$\%$. As the quantitative evaluation shown in Table~\ref{mainexpr}, our method excels in fault detection, with the prediction result closest to the ground truth, showing a smoother and more continuous effect of fault detection. In contrast, the prediction results of 2D models (UNet, HED, RCF, and DeeplabV3) have more noise, discontinuous fault detection, and fuzzy results. Two 3D models (FaultSeg Swin-UNET and Swin-UNETR) and the VIT and Swin Transformer models pre-trained on Imagenet tend to have interruptions in fault interpretation and contain some areas misjudged as faults and also have omissions. These results indicate that our method has apparent advantages in seismic fault detection.

\subsection{Ablation Study}

According to the ablation experiment results in Table~\ref{abl1}, we analyze the impact of different numbers of adjacent slices on model performance. The results show the best effect when using 5 adjacent slices as 2.5D input which indicates that this can effectively capture the inherent close volume correlation and achieve the best balance between model performance and computational efficiency. In addition, Table~\ref{abl2} further validates the effectiveness of data augmentation strategies. Specifically, the model performance is optimal when the horizontal invariance and wave propagation methods are combined. 
Finally, Table~\ref{abl3} compares the convergence speeds of different methods when training for just one epoch. Our method has a clear advantage in convergence speed; OIS and ODS reach 0.835 and 0.830, respectively, significantly higher than other models. This indicates that our method can learn the feature embeddings needed for fault segmentation in a short time, demonstrating excellent initial learning ability and efficient model optimization characteristics. This fast convergence capability is significant for rapid deployment and model updates in practical applications.

\begin{table}[htbp]
\caption{Ablation Study on the Number of Adjacent Slices.}
\label{abl1}
\centering
\setlength{\tabcolsep}{2.6pt} 
\begin{tabular}{cccc}
    \toprule
    \textbf{Number of Adjacent slices} & \textbf{OIS} & \textbf{ODS}\\
    \midrule
    M = 1 & 0.836 & 0.833 \\
    M = 3 & 0.862 & 0.858 \\
    M = 5 & 0.866 & 0.863 \\
    M = 7 & 0.858 & 0.855 \\
    \bottomrule
    \multicolumn{3}{l}{$^{\mathrm{a}}$Considering time cost, only 100 epochs were trained.}
\end{tabular}
\end{table}

\begin{table}[htbp]
\caption{Ablation Study on Data Augmentation.}
\label{abl2}
\centering
\setlength{\tabcolsep}{2pt} 
\begin{tabular}{cccc}
    \toprule
    \textbf{Data augmentation} & \textbf{OIS} & \textbf{ODS} \\
    \midrule
    Horizontal Invariance & 0.873 & 0.870 \\
    Wave Propagation & 0.867 & 0.870 \\
    All & 0.876 & 0.873 \\
    \bottomrule
    \multicolumn{2}{l}{$^{\mathrm{a}}$Considering time cost, only 100 epochs were trained.}
\end{tabular}
\end{table}

\begin{table}[htbp]
\caption{Evaluation of Model Convergence Speed Using 1 Epoch Training Results.}
\label{abl3}
\centering
\setlength{\tabcolsep}{2.6pt} 
\begin{tabular}{ccccc}
    \toprule
    \textbf{Model} & \textbf{Pre-trained method} & \textbf{OIS} & \textbf{ODS}\\
    \midrule
    ViT & w/o & 0.426 & 0.413 \\
    ViT & MAE & 0.385 & 0.374 \\
    ViT & Imagenet-pretrained & 0.647 & 0.643 \\
    Swin & w/o & 0.418 & 0.409 \\
    Swin & SimMIM & 0.694 & 0.689 \\
    Swin & Imagenet-pretrained & 0.731 & 0.726 \\
    2.5D SAM-h+Adapter+DA & SAM pre-trained & 0.835 & 0.830 \\
    \bottomrule
    \multicolumn{4}{l}{$^{\mathrm{a}}$w/o indicates without pre-training.}
\end{tabular}
\end{table}

\section{Conclusion}

This paper proposes Seismic Fault SAM, a lightweight Adapter method. It applies SAM to the seismic domain for the first time. We combine the 2.5D input design to capture 3D spatial patterns and use data augmentation methods based on priors to integrate geological constraints into the model. This approach successfully combines the general knowledge learned by large models with the fault detection task, achieving remarkable results. At present, we have only experimented with fault detection tasks. However, this method provides an effective extension plan for downstream tasks in the seismic field that lacks labeled data. The proposed Adapter method can be directly applied to various tasks and flexibly designed adaptation strategies, thereby unlocking greater potential for developing basic seismic models. In addition, we encourage more scholars to share their seismic data and annotations so that the model effect can be verified on more data sets. This also brings out our future research direction: introducing a human-in-the-loop approach, designing interactive systems to assist geologists in optimizing the fault interpretation process, and combining automated methods to obtain seismic benchmark datasets.

\bibliography{references}{}
\bibliographystyle{IEEEtran}

\end{document}